\documentclass[11pt,a4paper]{article}
\usepackage[hyperref]{acl2017}
\usepackage{times}
\usepackage{latexsym}
\usepackage{amsmath}
\usepackage{url}
\usepackage{todonotes}
\usepackage{graphicx}
\usepackage{array}
\usepackage{multirow}
\usepackage{amsfonts}

\aclfinalcopy 

\newcommand*\rot{\rotatebox[origin=c]{90}}
    \newcolumntype{C}[1]{>{\centering\let\newline\\\arraybackslash\hspace{0pt}}m{#1}}

\title{A Neural Architecture for Generating  Natural Language Descriptions from Source Code Changes}

\author{Pablo Loyola, Edison Marrese-Taylor and  Yutaka Matsuo \\
  Graduate School of Engineering \\
  The University of Tokyo \\
  Tokyo, Japan \\
  {\tt \{pablo,emarrese,matsuo\}@weblab.t.u-tokyo.ac.jp} }

\date{}

\begin{document}

\maketitle

\begin{abstract}
We propose a model to automatically describe changes introduced in the source code of a program using natural language. Our method receives as input a set of code commits, which contains both the modifications and  message introduced by an user. These two modalities are used to train  an encoder-decoder architecture. We evaluated our approach on twelve real world open source projects from four different programming languages. Quantitative and qualitative  results showed that the proposed approach can generate feasible and semantically sound descriptions not only in standard in-project settings, but also in a cross-project setting.

\end{abstract}

\section{Introduction}

Source code, while conceived as a set of structured and sequential instructions, inherently reflects human intent: it encodes the way we command a machine to perform a task. In that sense, it is expected that it follows to some extent the same distributional regularities that a proper natural language manifests \cite{hindle2012naturalness}. Moreover, the unambiguous nature of source code,  comprised in plain and  human-readable format,  allows an indirect way of communication between developers,  a phenomenon boosted in recent years given the current software development
paradigm, where billions of lines code are written in a distributed and  asynchronous way \cite{gousios2014exploratory}. 


The scale and complexity of software systems these days has naturally led to explore automated ways to support developers' code comprehension \cite{letovsky1987cognitive} from a linguistic perspective. One of these attempts is automatic summarization, which aims to generate a compact representation of the source code in a portion of natural language \cite{haiduc2010supporting}.


While existing code summarization methods are able to provide relevant insights about the purpose and functional features of the code, their scope is inherently static. In contrast, software development can be  seen as a sequence of incremental changes, intended to either generate a new functionality or to repair an existing one. Source code changes are critical for understanding program evolution, which motivated us to explore if it is possible to extend the 
notion of summarization to encode code changes into natural language representations, i.e., develop a model able to \emph{explain} a source code level modification. With this, we envision a tool for developers that is able to \emph{i)} ease the comprehension of the dynamics of the system, which could be useful for debugging and repairing purposes and \emph{ii)} automate the documentation of source code changes. 

To this end, we rely on the concept of code commit, the standard contribution procedure implemented in modern subversion systems \cite{gousios2014exploratory}, which provides both the actual change and a short explanatory paragraph. Our model consists of an encoder-decoder architecture which is trained on a set of triples conformed by the version of a system before and after the change, along with the comment. Given the high heterogeneity of the modalities involved, we rely on an attention mechanism to efficiently learn the parts of the sequences that are more expressive and  have more explanatory power.

We performed an empirical study on twelve real world software systems, from which we obtained the commit activity to evaluate our model. Our experiments  explored in-project and cross-project scenarios, and  our results showed that the proposed model is able to  generate semantically sound descriptions.

\section{Related Work}

The use natural language processing to support software engineering tasks has 
increased consistently over the years, mainly in terms of source code search, traceability and program feature location \cite{Panichella:2013:EUT:2486788.2486857,Asuncion:2010:STT:1806799.1806817}. 

The emergence of unifying  paradigms that explicitly relate programming and natural languages in distributional terms \cite{hindle2012naturalness} and the availability of large corpus mainly from open source software opened the door for the use of  language modeling for several tasks \cite{raychev2015predicting}. Examples of this are approaches for learning program representations \cite{mou2016convolutional}, bug localization \cite{huolearning}, API suggestion \cite{Gu:2016:DAL:2950290.2950334} and code completion \cite{raychev2014code}. 

Source code summarization has received special attention, ranging from the use of information retrieval techniques to the addition of physiological features such as eye tracking \cite{rodeghero2014improving}. In recent years several representation learning approaches have been proposed, such as \cite{allamanis2016convolutional}, where the authors employ a convolutional architecture embedded inside an attention mechanism to learn an efficient mapping between source code tokens and natural language keywords. 

More recently, \cite{iyer1summarizing} proposed a encoder-decoder model that learns to summarize  from Stackoverflow data, which contains snippet of code along with descriptions. Both approaches share the use of attention mechanisms \cite{bahdanau2014neural} to overcome the natural disparity between the  modalities when finding relevant token alignments. Although we also use an attention mechanism,  we differ from them in the sense we are targeting the changes in the code rather than the description of a file.

In terms of specifically working on code change summarization, \newcite{cortes2014automatically,linares2015changescribe} propose a method based on a set of rules that considers the type and impact of the changes, and  \cite{Buse:2010:ADP:1858996.1859005} combines summarization with symbolic execution. To the best of our knowledge, our approach represents the first attempt to generate natural language  descriptions from code changes without the use of hand-crafted features, a desirable setting given the heterogeneity of the data involved.

\section{Proposed Model}




Our model assumes the existence of  $T$ versions of a given project $\{v_1, \dots, v_T\}$.
Given a pair of consecutive versions $(v_{t-1}, v_{t})$, we define the tuple $(C_t,N_t$), where $C_t = \Delta_{t-1}^{t}(v)$ represents a code snippet associated to changes over $v$ in time $t$ and $N_t$ represents its corresponding natural language (NL) description. Let $\mathbb{C}$ be the set of all source code snippets and $\mathbb{N}$ be the set of all descriptions in NL. We consider a training corpus with $T$ code snippets and summary pairs $(C_t,N_t)$, $1 \leq t \leq T$, $C_t \in \mathbb{C}$ , $N_t \in \mathbb{N}$. Then, for a given code snippet $C_k \in \mathbb{C}$, the goal of our model is to produce the most likely NL description $N^{\star}$. 

Concretely, similarly to \cite{iyer1summarizing}, we use an attention-augmented encoder-decoder architecture. The encoder can be seen as a lookup layer, which simply reads through the source input sequence and returns the embedded tokens. The decoder is a RNN that reads this representation and generates NL words one at a time based on its current hidden state and guided by a global attention model \cite{luong-pham-manning:2015:EMNLP}. We model the probability of a description as a product of the conditional next-word probabilities. More formally, for each  NL token $n_i  \in  N_t$ we define,  
\begin{eqnarray}
h_i = f ( n_{i-1} E, h_{i-1} )\\
p(n_i | n_1, ... ,n_{i-1}) \propto W \tanh(W_1 h_i + W_2 a_i)
\end{eqnarray}

where $E$ is the embedding matrix for NL tokens, $\propto$ denotes a softmax operation, $h_i$ represents the hidden state and $a_i$ is the contribution from the attention model on the source code. $W$, $W_1$ and $W_2$ are trainable combination matrices. The decoder repeats the recurrence until a fixed number of words or a special \textit{END} token is generated. The attention contribution $a_i$ is defined as $a_i = \sum_{j=1}^k \alpha_{i,j} \cdot c_j F$, where  $c_j \in C_t$ is a source code token, $F$ is the source code token embedding matrix and  $\alpha_{i,j}$ is: 
\begin{eqnarray}
\alpha_{i,j} = \frac{\exp{(h_i^{\top} c_j F)}}{\sum_{c_j \in C_t} \exp{(h_i^{\top} c_j F)}}
\end{eqnarray}

We use a dropout-regularized LSTM cell for the decoder  \cite{zaremba2014} and also add dropout at the NL embeddings and at the output softmax layer, to prevent over-fitting. We added special \textit{START} and \textit{END} tokens to our training sequences and replaced all tokens and output words occurring less than 2 and 3 times, respectively, with a special \textit{UNK} token. We set the maximum code and NL length to be 100 tokens. For decoding, we approximate $N^{\star}$ by performing a beam search on the space of all possible summaries using the model output, with a beam size of 10 and a maximum summary length of 20 words.

To evaluate the quality of our generated descriptions we use both METEOR \cite{Lavie:2007:MAM:1626355.1626389} and sentence level BLEU-4 \cite{papineni-EtAl:2002:ACL}. Since the training objective does not directly optimize for these scores, we compute METEOR on our validation set after every epoch and save the intermediate model that gives the maximum score as the final model. For evaluation on our test set we used the BLEU-4 score. 


\section{Empirical Study}

\textbf{Data and pre-processing:} We captured historical data from twelve open source projects hosted on Github based on their popularity and maturity, selecting 3 projects for each of the following languages: \emph{python}, \emph{java}, \emph{javascript} and \emph{c++}. For each project, we downloaded diff files and metadata of the full commit history. Diff files encode per-line differences between two files or sets of files in a standard format, allowing us to recover source code changes in each commit at the line level. On the other hand, medatada allows us to recover information such as the author and message of each commit.


The extracted  commit messages were processed using the Penn Treebank tokenizer \cite{marcus1993building}, which nicely deals with punctuation and other text marks. To obtain a source code representation of each commit, we parsed the diff files and used a lexer \cite{lexer} to tokenize their contents in a per-line fashion  allowing us to maximize the amount of source code recovered from the diff files.  Data and source code available\footnote{\url{http://github.com/epochx/commitgen}}.


\textbf{Experimental Setup:} Given the flat structure of the diff file, source code in contiguous lines might not necessarily correspond to originally neighboring code lines. Moreover, they might come from different files in the project. To deal with this issue, we first worked only with those commits that modify a single file in the project; we call this the \textit{atomicity} assumption. By using only \textit{atomic} commits we reduced our training data by an average of roughly 50\%, but in exchange we made sure all the extracted code lines came from the same file. At the same time, we expect to  maximize the likelihood of observing a direct relation between the commit message and the lines altered.

We then relaxed our \textit{atomicity} assumption and experimented with the \textit{full} commit history. Given our maximum sequence length constrain of 100 tokens, we only observed an average of 1,97\% extra data on each project. Since source code lines may come from different files, we added a delimiting token \textit{NEW\_FILE} when corresponding.

We were also interested in studying the performance of the model in a cross-project setting. Given the additional challenges that this involves, we designed a more controlled experiment. Starting from the  \textit{atomic}
dataset, we selected commits that only add or only remove code lines, conforming a derived dataset that we call \textit{uni-action}. We chose the  \textit{python} language to maximize the available data. See Table \ref{table:data_summary}.


\begin{table}
    \centering
    \scriptsize
    \begin{tabular}[h]{ c c c c c c }
        \hline
        \textbf{Language}	&	\textbf{Project} &	\textbf{Full}	& \textbf{Atomic}	& \textbf{Added}    &\textbf{Rem}. \\ \hline
        \multirow{3}{*}{python} 
        			&	Theano			&	24,200	& 65.40\%	& 11.43\%	& 2,83\% 	\\ 
                    &	keras			&	2,855	& 66.02\%	& 11.07\%	& 3,01\%	\\ 
                    &	youtube-dl		&	13,968	& 74.49\%	& 11.52\%	& 2,59\%	\\ \hline
        \multirow{3}{*}{javascript}
        			&	node			&	15,811	& 53.17\%	& 11.87\%	& 3,21\%	\\ 
        			&	angular			&	6,204	& 32.90\%	& 5.59\%	& 1,72\%	\\ 
        			&	react			&	7,806	& 53.29\%	& 12.67\%	& 2,72\%	\\ \hline
        \multirow{3}{*}{c++}
        			&	opencv			&	20,480	& 50.08\%	& 8.83\%	& 1,66\%	\\ 
        			&	CNTK			&	10,792	& 38.36\%	& 6.00\%	& 2,23\%	\\ 
        			&	bitcoin			&	12,596	& 48.11\%	& 9.84\%	& 2,56\%	\\ \hline
        \multirow{3}{*}{java}
        			&	CoreNLP			&	9,149	& 42.77\%	& 7.84\%	& 1,98\%	\\ 
        			&	elasticsearch	&	25,764	& 43.77\%	& 9.02\%	& 2,61\%	\\ 
        			&	guava			&	3,821	& 38.63\%	& 8.90\%	& 2,64\%	\\ \hline
        \multicolumn{2}{c}{Average}	& 	12,787	& 50.58\% 	& 9.55\%	& 2,48\%	\\ \hline
    \end{tabular}
    \caption{Summary of our collected data.}
    \label{table:data_summary}
\end{table}

\textbf{Results and Discussion:} We begin by training our model on the \textit{atomic} dataset. As baseline we used MOSES \cite{koehn-EtAl:2007:PosterDemo} which although is designed as a phrase-based machine translation system, was previously used by \newcite{iyer1summarizing} to generate text from source code. Concretely, we treated the tokenized code snippet as the source language and the NL description as the target. We trained a 3-gram language model using KenLM \cite{Heafield-estimate} and used mGiza to obtain alignments. For validation, we use minimum error rate training \cite{bertoldi2009,och:2003:ACL} in our validation set.

As Table \ref{table:atomic_full_results} shows, our model trained on \textit{atomic} data outperforms the baseline in all but one project with an average gain of 5 BLEU points. In particular, we observe bigger gains for java projects such as \textit{CoreNLP} and \textit{guava}. We hypothesize this is because program differences in Java tend to be longer than the rest. While this impacts on training time, at the same time it allows the model to work with a larger vocabulary space. On the other hand, our model performs similarly to MOSES for the \textit{node}  and slightly worse for the \textit{youtube-dl}. A detailed inspection of the NL messages for \textit{node} showed that many of them exhibit a fixed pattern in their structure. We believe this rigidity restrains the generation capabilities of the decoder, making it more prone to memorization. 

Table \ref{table:gen_examples} shows examples of generated descriptions for real changes and their references. Results  suggest  that our model is able to generate semantically sound  descriptions for the changes. We can also visualize the summarizing power of the model, as seen in the \textit{Theano}  and  \textit{bitcoin} examples. We observe a tendency to choose more general terms over too specific ones meanwhile also avoiding  irrelevant words such as numbers or names. Results also suggest the emergence of rephrasing capabilities, specifically in the second  example from \textit{Theano}. Finally, our generated descriptions are, in most cases, semantically well correlated to the reference descriptions. 
We also report  not so successful results, such as  case of \textit{youtube-dl}, where we can see signs of memorization on the generated descriptions. 

Regarding the cross-project setting experiments on \textit{python}, we obtained BLEU scores of 14.6 and 18.9 for only-adding and only-removing instances in the \textit{uni-action} dataset, respectively. We also obtained validation accuracies up to 43.94\%, suggesting feasibility in this more challenging  scenario. Moreover, as the generated descriptions from the \textit{keras} project in Table \ref{table:gen_examples} show, the model is still able to generate semantically sound descriptions.


\begin{figure}[h]
    \includegraphics[width=0.25\textwidth]{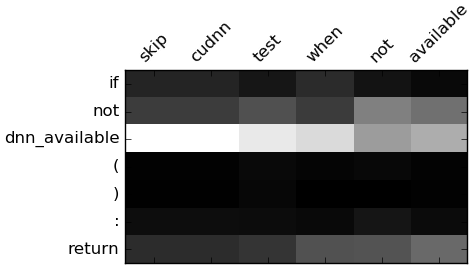}
    \includegraphics[width=0.22\textwidth]{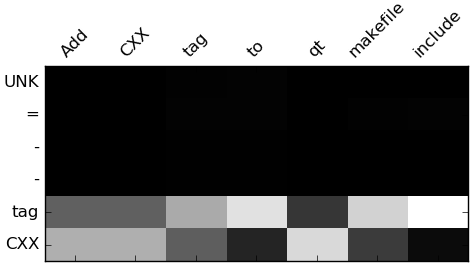}
    \caption{Heatmaps of attention weights $\alpha_{i,j}$.}
    \label{fig:attention}
\end{figure}

\begin{table}[t]
    \centering
    \scriptsize
    \begin{tabular}{ |c|C{0.2\textwidth}|C{0.2\textwidth}| }
        \hline
         & \textbf{Reference} & \textbf{Generated} \\ \hline
        \multirow{2}{*}{\rot{keras}}    & Fix image resizing in preprocessing/image & Fixed image preprocessing . \\ \cline{2-3}
                                        & Fix test flakes & Fix flaky test \\ \hline
        \multirow{3}{*}{\rot{Theano}}   & fix crash in the new warning message .    & Better warning message . \\ \cline{2-3}
                                        & remove var not used . & remove not used code . \\ \cline{2-3}
                                        & Better error msg & better error message . \\ \hline
        \multirow{2}{*}{\rot{bitcoin}}  & Merge pull request  4486 45abeb2 Update Debian packaging description for new bitcoin-cli ( Johnathan Corgan ) & Update Debian packaging description for new bitcoin-cli \\ \cline{2-3}
                                        & Add two unittest-related files to .gitignore  &   Add : Minor files to .gitignore \\ \hline
        \multirow{2}{*}{\rot{CoreNLP}}  & Add a bunch of verbs which are more likely to be xcomp than vmod & Add a bunch of verbs which are more to be xcomp than vmod \\ \cline{2-3}
                                        & Add a brief test for optional nodes   & make this test do something \\ \hline
        \multirow{3}{*}{\rot{youtube-dl}} & [ crunchyroll ] Fix uploader and upload date extraction & [ crunchyroll ] Fix uploader extraction \\ \cline{2-3}
                                        & [ extractor/common ] Improve base url construction & [ extractor/common ] Improve extraction \\ \cline{2-3} 
                                        & [ mixcloud ] Use unicode\_literals & [ common ] Use unicode\_literals \\ \hline
        \multirow{2}{*}{\rot{opencv}}   & fixed gcc compilation & fixed compile under linux \\ \cline{2-3}
                                        & remove unused variables in OCL\_PERF\_TEST\_P ( ) & remove unused variable in the module \\ \hline
    \end{tabular}
    \caption{Examples of generated natural language passages v/s original ones taken from the test set.}
    \label{table:gen_examples}
\end{table}

Despite the small data increase, we also trained our model on \textit{full} datasets as a way to confirm the generative power of our model. In particular, we wanted to test the model is able leverage on \textit{atomic} data to also capture and compress multi-file changes. As shown in Table \ref{table:atomic_full_results}, results in terms of BLEU and validation accuracy manifest reasonable consistency, despite the higher disparity between source code and natural language on this dataset, which means the model was able to learn representations with more compressive power.


 Soft alignments  derived from Figure \ref{fig:attention}, which shows examples of attention heatmaps, illustrate how the model effectively associates source code tokens with meaningful words.

\begin{table}[h]
    \scriptsize
    \centering
    \begin{tabular}{c|c c c |c c c}
        \hline
        \multirow{2}{*}{\textbf{Dataset}} & \multicolumn{3}{c}{\textit{atomic}} & \multicolumn{2}{c}{\textit{full}} \\ \cline{2-6}
         & \textbf{Val. acc} & \textbf{BLEU} & \textbf{Moses}    &\textbf{Val. acc}     & \textbf{BLEU} \\ \hline
        Theano              & 36.81\%       & 9.5           & 7.1 	            & 39.88\%          & 10.9 \\ 
        keras               & 45.76\%       & 13.7          & 7.8 	            & 59.30\%          & 8.8 \\ 
        youtube-dl          & 50.84\%       & 16.4          & 17.5              & 53.65\%          & 17.7 \\ 
        node                & 52.46\%       & 7.8           & 7.7 	            & 53.70\%          & 7.2 \\ 
        angular             & 44.39\%       & 13.9          & 11.7              & 45.06\%          & 15.3 \\ 
        react               & 49.44\%       & 11.4          & 10.7              & 48.61\%          & 12.1  \\ 
        opencv              & 50.77\%       & 11.2          & 9.0 	            & 49.00\%          & 8.4 \\ 
        CNTK                & 48.88\%       & 17.9          & 11.8              & 44.85\%          & 9.3  \\ 
        bitcoin             & 50.04\%       & 17.9          & 13.0              & 55.03\%          & 15.1 \\ 
        CoreNLP             & 63.20\%       & 28.5          & 10.1              & 62.25\%          & 26.7 \\ 
        elasticsearch       & 36.53\%       & 11.8          & 5.2	            & 35.98\%          & 6.4 \\ 
        guava               & 65.52\%       & 29.8          & 19.5              & 67.15\%          & 34.3 \\ \hline
    \end{tabular}
\caption{Results on the \textit{atomic} and \textit{full} datasets.}
\label{table:atomic_full_results}
\end{table}







\section{Conclusion and  Future work}
We proposed an encoder-decoder model  for automatically generating natural descriptions from source code changes. 
We believe our current results suggest that the idea is feasible and, if improved, could represent a contribution for the understanding of software evolution from a linguistic perspective.  As future work,  we will consider improving the model by allowing feature learning from richer inputs, such as abstract syntax trees and  also functional data, such as execution traces.



\newpage

\bibliography{acl2017}
\bibliographystyle{acl_natbib}

\end{document}